\documentclass{article}
\pdfoutput=1
\usepackage[final,nonatbib]{nips_2017}

\usepackage[utf8]{inputenc} 
\usepackage[T1]{fontenc}    
\usepackage{hyperref}       
\usepackage{url}            
\usepackage{booktabs}       
\usepackage{amsfonts}       
\usepackage{nicefrac}       
\usepackage{microtype}      

\usepackage[pdftex]{graphicx} 

\usepackage{subfigure}

\title{Towards Deep Learning Models for Psychological State Prediction using Smartphone Data:\\ Challenges and Opportunities}

\author{
  Gatis Mikelsons\\
  University of Oxford and \\ The Alan Turing Institute\\
  \texttt{gatis.mikelsons@jesus.ox.ac.uk} \\
  \And
  Matthew Smith \\
  University of Warwick and \\ The Alan Turing Institute\\
  \texttt{m.d.smith@warwick.ac.uk} \\
  \AND
  Abhinav Mehrotra \\
  University College London and \\ The Alan Turing Institute \\
  \texttt{a.mehrotra@ucl.ac.uk} \\
  \And
  Mirco Musolesi \\
  University College London and \\ The Alan Turing Institute \\
  \texttt{m.musolesi@ucl.ac.uk} \\
}

\begin{document}

\maketitle

\begin{abstract}


There is an increasing interest in exploiting mobile sensing technologies and machine learning techniques for mental health monitoring and intervention. Researchers have effectively used contextual information, such as mobility, communication and mobile phone usage patterns for quantifying individuals' mood and wellbeing. In this paper, we investigate the effectiveness of neural network models for predicting users' level of stress by using the location information collected by smartphones. We characterize the mobility patterns of individuals using the GPS metrics presented in the literature and employ these metrics as input to the network. We evaluate our approach on the open-source StudentLife dataset. Moreover, we discuss the challenges and trade-offs involved in building machine learning models for digital mental health and highlight potential future work in this direction.

\end{abstract}

\section{Introduction}

Mobile phones have transformed over a period of time from merely communication tools to an indispensable part of daily life assisting us in a variety of day-to-day situations. At the same time, these devices come with an array of embedded sensors that are capable of passively monitoring numerous physical-context modalities. However, these sensors are not capable of directly capturing users' cognitive context, such as mood and well-being states. Addressing this challenge by enabling mobile phones to passively infer them could help tackle the global burden caused by adverse mental health conditions~\cite{pejovic2017anticipation}. Moreover, early detection of such conditions is essential for effective prevention by enabling appropriate intervention and treatments. 

The new technological capability of obtaining an unprecedented access to people's emotional states and aspects of daily lives has given rise to a growing number of mental-health apps~\cite{anthes2016pocket}, and the task of making sense of such data and building prediction models is one of great interest to researchers and practitioners~\cite{moutoussis2017building}. Recent studies have shown the potential of exploiting mobile sensing data to learn and, potentially, predict the users' cognitive context, such as mood~\cite{Emotionsense2010, Likamwa2013MoodScope, Servia2017Mobile, Mehrotra2017MyTraces} and well-being states~\cite{Canzian2015MoodTraces, Saeb2015Depression, Mehrotra2016MultiModal, Suhara2017DeepMood}. Other recent studies have focused on the analysis of social media information~\cite{benton2017multitask,gkotsis2017characterisation, amir2017quantifying}. However, until now, very few of these studies have explored the potential of neural-network-based machine learning algorithms to model users' mood through their contextual data from mobile phones. Solely, Suhara et al.~\cite{Suhara2017DeepMood} have investigated the development of a neural-network-based algorithm for inferring bipolar states of a user through behavioral self-reports collected via mobile phones. In particular, the authors developed a bipolar depression forecasting model based on long short-term memory recurrent neural networks (LSTM-RNNs)~\cite{LSTM}. The app does not exploit sensor data for depression inference.

In this preliminary study, we investigate the potential of a deep-learning approach~\cite{lecun2015deep} for the design of classifiers of psychological states based on features derived from GPS mobility patterns of individuals. We focus on stress, but the methodology and techniques described here can be generalized to other conditions. To evaluate our model, we use the StudentLife dataset~\cite{SL}, which contains rich and multi-dimensional data collected from students at Dartmouth College over the course of a term. We consider eight GPS metrics, based on the recent literature~\cite{Mehrotra2017Designing}, and four temporal variables to indicate weekends and the start, middle and end of the term. We evaluate the model using the cross-validation approach and present results comparing the importance of these two feature types. Finally, we discuss challenges in this area, potential future approaches and pitfalls in effective variable selection.

\section{Approach}

We rely on the Experience Sampling Method~\cite{Csikszentmihalyi1983ESM} to collect users' stress states, which we use as ground truth. Most mood questionnaires are reported on a Likert scale, which could be perceived differently by users, e.g., slightly stressed for a user could be a response of 4 on the given 1-5 scale, while a similar level of stress could be reported as 5 by another user. Indeed, the classification task is particularly hard in case of psychological states given the fact that there is an inherent "noise" associated with the labels. To address this issue, we apply a technique of daily averaging and rescaling of the data into three stress levels: below-median, median and above-median. Here, median is computed for each individual separately for their entire data. 

\begin{figure}[h]
\centering
\includegraphics[width=10cm]{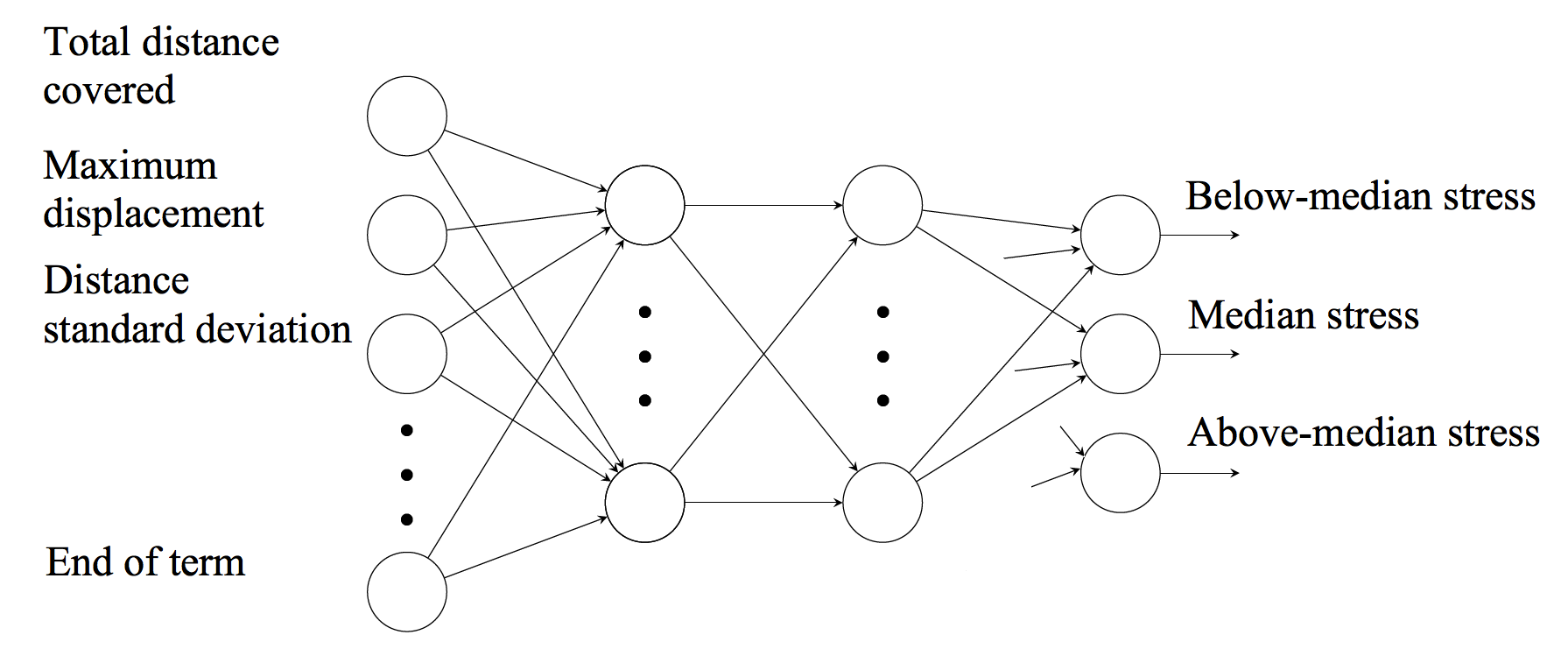}\\
\caption{Four-layer fully connected network for a three-class classification task.}
\label{figx}
\end{figure}

We construct a neural network model and train it with the aim of predicting these stress reports based on predictor variables calculated for the same day. Eight spatio-temporal metrics based on the recent literature~\cite{Mehrotra2017Designing} are used as predictors: total distance covered in a day, maximum 2-point displacement in a day, distance standard deviation, number of different areas visited by tiles approximation, total spatial coverage by convex hull, difference in sequence of tiles covered compared to previous day, difference in sequence of clusters visited compared to previous day, distance entropy. After calculating the metrics based on the GPS data, they are standardised on per-user basis. Moreover, we add four temporal one-hot features indicating weekends, start of term, midterm, end of term. This increases the number of input nodes to 12. The data set contains information that can be used for construction of further features (e.g., deadlines, activity data, call information), which we are currently investigating.

The use of a neural network predictor is motivated, in part, by the absence of previous work in the field using this model type. Neural networks are known for their ability to learn non-linear dependencies and discern subtle high-level features, which may be essential qualities for successful modeling of human mental states. We report the results of a network that uses four fully connected layers (sizes: 57, 35, 35, 3). The first three layers feature $tanh$ activation, and we apply batch normalization for each as well as drop-out regularization of rates 0.35, 0.25, and 0.15. The final layer is a $softmax$-activation layer with batch normalization applied and three output nodes corresponding to the three stress classes. We find empirically that using such a model of considerable complexity together with heavy regularisation yields the best performance on the data set. As a further measure, early stopping based on validation loss is applied in order to control against overfitting (see Figure \ref{fig2} left). Training is done on the categorical cross-entropy score using the Adam optimizer~\cite{kingma2014adam}. We employ a stratified 5-fold split, and we use the F1 score as the primary performance metric, given the peaked distribution of classes in our data set. Precision and recall are also reported.

\section{Preliminary study}

\subsection{Description of the Dataset}

The publicly available StudentLife dataset \cite{SL} is an ambitious study carried out at Dartmouth College during the spring term of 2013, which gathered information about 49 students, undergraduates and graduates. It comprises a vast complex of GPS traces, conversation times, phone lock information, psychological pre and post-term assessment data, GPA information and accelerometer readings, among other measurements. Interesting machine-learning studies have been carried out using the dataset, including a GPA prediction study \cite{smartgpa}. Students were also subjected to "ecological momentary assessments" (EMAs) whereby the phone would prompt them to answer short in-situ surveys about their health and mood, about aspects of college life, or just to submit a random comment. Such EMAs, typically being sent to students more than once a day, create a trace of real-time self-reported information in the dataset, which is of particular interest from the modeler's perspective.

\begin{figure}[t]
	\centering
	\begin{tabular}{cc}
	\includegraphics[width = 0.6\textwidth]{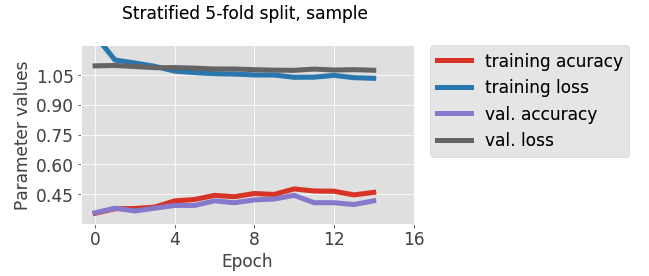} &
	\includegraphics[width = 0.35\textwidth]{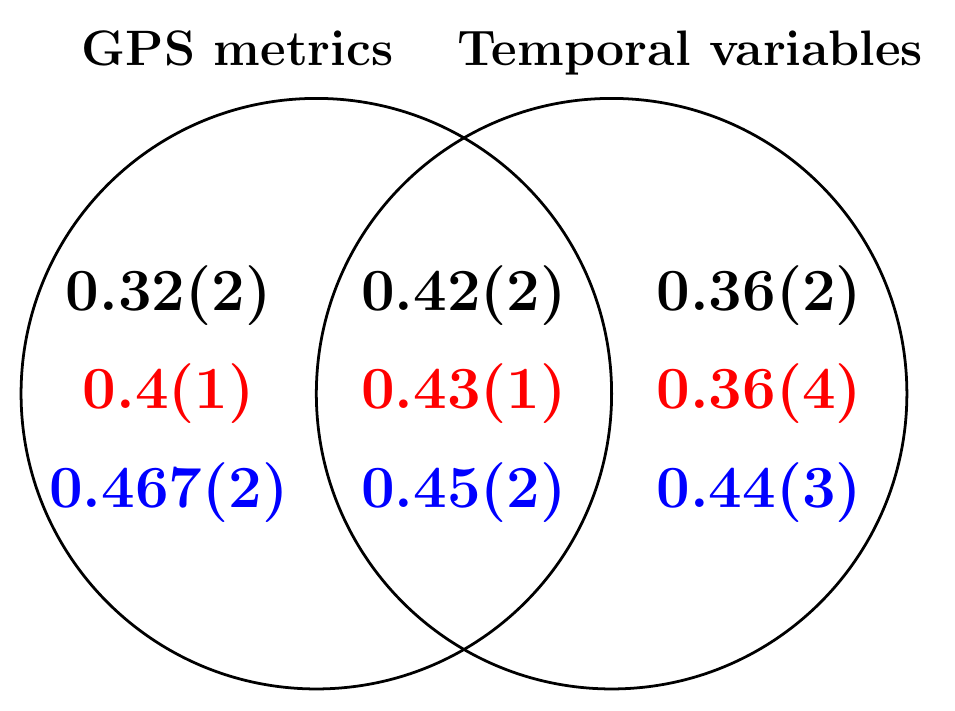} 
	\end{tabular}
	\caption{Left: sample run of the prediction model with the 12 features included. Right: figures for the F1 score (black), precision (red) and recall (blue) using 3 feature combinations (GPS metrics, temporal variables and all the features).}
	\label{fig2}
\end{figure}

We consider the results for the Stress EMA. Students would be prompted by their phones to answer the inquiry: "Right now, I am \dots" with one of five choices: "A little stressed", "Definitely stressed", "Stressed out", "Feeling good", "Feeling great". In total, approximately 2300 momentary responses are recorded. The number of responses per student range from 1 to 269, and response times are distributed across all the waking hours. After daily averaging and matching, where available, with the corresponding GPS metrics, we obtain a set of 1078 stress responses and corresponding features. The distribution of stress responses is peaked, with 47\% reports of "median" stress. 

\subsection{Initial Results}

The left panel in Figure \ref{fig2} shows the parameter evolution for a sample fold with all 12 input features, where early stopping is applied. The Venn diagram in the right panel provides an insight into how the two feature types compare in predictive power. We report results for GPS metrics and temporal variables each used separately as well as combined. The average F1 score (black), precision (red) and recall (blue) are displayed, where we quote the single standard deviation error from 5-fold stratified validation. The metrics for each run are computed by weighted averaging, based on the number of true instances for each label.

For comparison, the F1 score obtained by guessing the most frequent category is 0.30 in all our stratified samples. Similarly, we obtain precision of 0.22 and recall of 0.47. Using all the 12 input features, we obtain an F1 score that beats the mode classifier baseline. This result is encouraging, given the modest breadth of input features used and the complexity of the problem. The right panel in Figure \ref{fig2} provides a way of comparing the relative predictive power of the temporal and the GPS metrics.

In future work, we intend to extend the range of features used in the model as well as compare against more sophisticated baselines, such as the random forest classifier. This would also provide another way of assessing the relative importance of various features.

\section{Challenges and Outlook}

Several general challenges will have to be addressed in order to implement robust solutions for successfully monitoring and predicting human mental states in real-world settings. In this section, we outline some key aspects that are of fundamental importance for realizing this vision.

\textbf{Establishing Ground Truth: Recall Bias and Biophysical Sensors.} An initial problem is to choose the right time horizon, i.e. the number of hours or days over which an instance of mood is defined and features calculated. In order to have an effective training set, it might be possible to introduce an explicit time frame in the question formulation ("Over the past x hours \dots"), which would remedy the issue to some extent. However, it can introduce recall bias, as users might not necessarily recall their past experiences accurately. On the other hand, one can also ask whether self-reported mood is the best approach for establishing ground truth. Other measurements, such as galvanic skin response, have been studied~\cite{sano2013stress} and could be considered. Indeed, in the case of psychological classification tasks, labels might be much more "noisy" than for other typical deep learning applications.

\textbf{Fine-grained and Informative Features.} Another challenge is to abstract raw sensor data into an effective set of predictors. The present work has focused on hand-crafted and interpretable features. However, it is possible that there are more informative signals in the data that we are missing. To address this challenge, one can envisage other approaches, such as training a neural-network-based autoencoder for feature extraction. Another approach could be to manually codify the contextual information from sensor data. However, this would be potentially very expensive and difficult to generalize. This is indeed an open area for this field. Furthermore, a great variety of information could be used to build features. For example, some studies~\cite{jaques2017predicting} consider weather in their models. It may also be effective to add participants' gender, their birthdays and paydays, time spent with family and at work, among many other possible contributing factors. 

\textbf{Heterogeneous Population.} A potential hindrance for building generalizable mood prediction models (i.e., one model for all) may be provided by human heterogeneity. Different predictors may cause mental effects of different direction and strength for different individuals even in a rather homogeneous pool. This would easily become a source of "noise" in a general model, suggesting that only modest accuracy at test time is attainable, frustrating the task of comparing different model designs against each other. Moreover, within a heterogeneous population, even if a single subject displays strong regularities, such as being stressed almost every Tuesday, they would be averaged into invisibility by training a general model. This suggests individual tailoring as a possible way forward, or clever approaches to clustering the participants into distinct groups. Domain adaptation, where general features and individual tailoring both play a role, using an approach similar to that presented in \cite{jaques2017predicting}, could also be a potentially fruitful approach.

\textbf{Large-scale and High-quality Data Collection.} There could be some drawbacks in surveying  human subjects engaged in their day-to-day lives. The gradual loss of users' interest and interrupting them at inconvenient times are common issues. Such intrusive nature of data collection might adversely impact data quality by leading users to answer falsely, not respond at all or even leave the study. These issues are a cause of drop in data quality as well as its amount. Moreover, relying on volunteers for data collection also risks introducing unexpected selection effects in the models built.

\section{Conclusion}

We have developed and presented a deep-learning model for stress prediction based on the StudentLife dataset. Possible approaches to performance analysis and variable inspection are suggested. We outline the main research challenges and opportunities in this area, such as listing and formalising effective predictor variables and dealing with population heterogeneity. We plan to investigate these issues using different and larger datasets, also considering more feature types and other baselines.

\subsubsection*{Acknowledgments}
This work was supported by The Alan Turing Institute under the EPSRC grant EP/N510129/1. We would like to thank the authors of the Studentlife dataset for making it available for the research community.

\bibliographystyle{abbrv}
\bibliography{paper}

\end{document}